\documentclass{article} 
\usepackage[preprint]{colm2026_conference}

\usepackage{microtype}
\usepackage{hyperref}
\usepackage{url}
\usepackage{booktabs}
\usepackage{amsmath}
\usepackage{times}
\usepackage{enumitem}
\usepackage{multirow}
\usepackage{xurl}
\usepackage{latexsym}
\usepackage{booktabs}
\usepackage{amsmath, amssymb}
\usepackage{graphicx}

\usepackage{bigstrut}
\usepackage{bbm}
\usepackage{bbding}
\usepackage{bbold}
\usepackage{mathtools}
\usepackage{xcolor}
\usepackage[table]{xcolor}
\usepackage{tcolorbox}
\usepackage{mathptmx}
\usepackage{hyperref}
\usepackage{amsthm}
\usepackage{subcaption}
\usepackage{tabularx}
\usepackage{mdframed}
\usepackage{wrapfig}

\newmdenv[
  leftline=true,
  rightline=false,
  topline=false,
  bottomline=false,
  linewidth=1pt,
  linecolor=black,
  skipabove=8pt,
  skipbelow=8pt,
  innerleftmargin=10pt,
  innerrightmargin=10pt
]{leftbarquote}

\usepackage{lineno}
\usepackage{inconsolata}
\usepackage{xspace} 

\definecolor{darkblue}{rgb}{0, 0, 0.5}
\hypersetup{colorlinks=true, citecolor=darkblue, linkcolor=darkblue, urlcolor=darkblue}

\newcommand {\bench} {{\textsc{PreScam}}\xspace}

\NewDocumentCommand{\weixiang}
{ mO{} }{\textcolor{orange}{\textsuperscript{\textit{Weixiang}}\textsf{\textbf{\small[#1]}}}}

\NewDocumentCommand{\shang}
{ mO{} }{\textcolor{red}{\textsuperscript{\textit{Shang}}\textsf{\textbf{\small[#1]}}}}

\title{PreScam: A Benchmark for Predicting Scam Progression from Early Conversations}

\author{Weixiang Sun\thanks{Equal contribution. Email: \texttt{\{wsun4,sma5\}@nd.edu}}, 
Shang Ma\footnotemark[1], 
Yiyang Li, Tianyi Ma, Zehong Wang, Colby Nelson \\
University of Notre Dame\\
\And
Xusheng Xiao \\
Arizona State University \\
\And
Yanfang Ye\thanks{Corresponding author. Email: \texttt{yye7@nd.edu}} \\
University of Notre Dame
}

\begin{document}

\ifcolmsubmission
\linenumbers
\fi

\maketitle

\begin{abstract}
Conversational scams, such as romance and investment scams, are emerging as a major form of online fraud. Unlike one-shot scam lures such as fake lottery or unpaid toll messages, they unfold through multi-turn conversations in which scammers gradually manipulate victims using evolving psychological techniques. However, existing research mainly focuses on static scam detection or synthetic scams, leaving open whether language models can understand how real-world scams progress over time. We introduce \bench, a benchmark for modeling scam progression from early conversations. Built from user-submitted scam reports, \bench filters and structures 177,989 raw reports into 11,573 conversational scam instances spanning 20 scam categories. Each instance is hierarchically structured according to the scam lifecycle defined by the proposed scam kill chain, and further annotated at the turn level with scammer psychological actions and victim responses.
We benchmark models on two tasks: real-time termination prediction, which estimates whether a conversation is approaching the termination stage, and scammer action prediction, which forecasts the scammer's subsequent actions. Results show a clear gap between surface-level fluency and progression modeling: supervised encoders substantially outperform zero-shot LLMs on real-time termination prediction, while next-action prediction remains only moderately successful even for strong LLMs. Taken together, these results show that current models can capture some scam-related cues, yet still struggle to track how risk escalates and how manipulation unfolds across turns. 

\end{abstract}









    

        

\section{Introduction}
Scams have long accompanied the evolution of human society and communication technology. Fraudulent schemes have historically adapted to the dominant communication medium of each era, from word-of-mouth deception in ancient societies~\citep{dill2022greekfraud,sun2023chinesefraud}, to print-based fraud such as the Spanish Prisoner~\citep{wikiadvancefee}, to telephone-based schemes such as boiler-room and Ponzi-style operations~\citep{secboilerroom}. Today, the Internet and social media have enabled scams that are more scalable, personalized, and psychologically sophisticated than ever before. Many modern scams, including investment, employment, and romance scams, do not unfold as isolated messages; instead, they develop through multi-turn interactions in which scammers gradually manipulate victims over time. In this work, we refer to such engagement-heavy scams as \textbf{conversational scams}.  
According to the Better Business Bureau (BBB) Annual Scam Risk Reports (2016–2024), major conversational scams (e.g., investment and employment scams) have emerged to dominate the riskiest scams over the last decade~\citep{bbbanualreport}.

A defining characteristic of conversational scams is that they are \emph{processes} rather than static artifacts. Scammers typically begin by establishing contact, then sustain engagement through repeated psychological manipulation, and finally attempt to extract money, credentials, or other assets. This progression is often strategic rather than random. Prior work in psychology and scam studies suggests that scammers rely on recurring psychological techniques such as authority, urgency, trust building, and fear induction, to shape victims' decisions over time~\citep{lea2009psychology,ma-etal-2025-psyscam,huang2024social, wang2026reasoning}. This observation raises an important modeling question: beyond recognizing isolated scam cues, can language models understand \emph{how} scams unfold as sequences of psychological manipulation?

Despite growing interest in LLMs for fraud analysis, existing efforts mostly study scams through either static detection or synthetic simulation~\citep{eder2025honeypot,yang2025fraud,ma2025teleantifraud,kumarage2025personalized,ye2025llms4all}. While promising, these approaches face two major limitations. First, synthetic conversations are often shaped by the inductive biases of the generating model and may not faithfully reflect the diversity and messiness of real-world scam reports. Second, most existing formulations treat scam conversations as unstructured text, overlooking the latent sequential structure that governs how scammers escalate manipulation across turns. As a result, current benchmarks provide limited insight into whether models truly capture the dynamics of scam progression.


To address this gap, we collect a large corpus of real-world scam reports from a prominent scam-reporting platform and transform them into 11,573 structured multi-turn scam conversations. Specifically, we introduce a new representation, termed the \textit{Scam Kill Chain}, inspired by prior work in cybersecurity and cognitive science~\citep{attck,montanez2022cyber,lea2009psychology,ma-etal-2025-psyscam} that formalizes cyber attacks and social engineering as structured, staged processes. The Scam Kill Chain formalizes each scam conversation through three lifecycle stages: \textit{Initial Contact}, \textit{Engagement}, and \textit{Termination}, and represents each stage using scammer actions grounded in psychological techniques, which we refer to as \textbf{PT Actions}. By explicitly modeling both the temporal phase of the scam and the underlying psychological manipulation, this representation converts raw scam narratives into structured manipulation trajectories and enables scam progression to be studied as a sequential reasoning problem.

Based on this representation, we present \bench, the first benchmark for modeling scam progression from early conversations. \bench is designed to evaluate whether models can track the evolving risk of an interaction and anticipate the scammer's next move from partial context. Concretely, we consider two tasks: \textbf{real-time termination prediction}, which measures whether a model can identify when an ongoing interaction is approaching the termination phase, and \textbf{scammer action prediction}, which tests whether a model can predict the scammer's actions conditioned on the observed history. Together, these tasks move beyond conventional scam classification and directly probe whether models can recover the underlying progression structure shaped by psychological manipulation.

Our study yields three main contributions. 

\begin{itemize}[noitemsep, topsep=1pt, partopsep=1pt, listparindent=\parindent, leftmargin=*]
    \item First, we construct a large-scale benchmark of real-world conversational scams with structured stage annotations and turn-level psychological technique labels. 
    \item Second, we provide a quantitative characterization of scam progression, revealing recurring stage-specific and scam-type-specific manipulation patterns.
    \item Third, we systematically evaluate a range of language models and baselines on scam progression modeling. Our results show that while current models capture useful scam-related priors, they remain limited in modeling the sequential, psychologically grounded actions that drive scam escalation.
\end{itemize}
We hope \bench\ can serve as a useful community benchmark for evaluating whether models can track risk and forecast scammer actions in real-world scam conversations.

\begin{figure}
    \centering
    \includegraphics[width=0.99\linewidth]{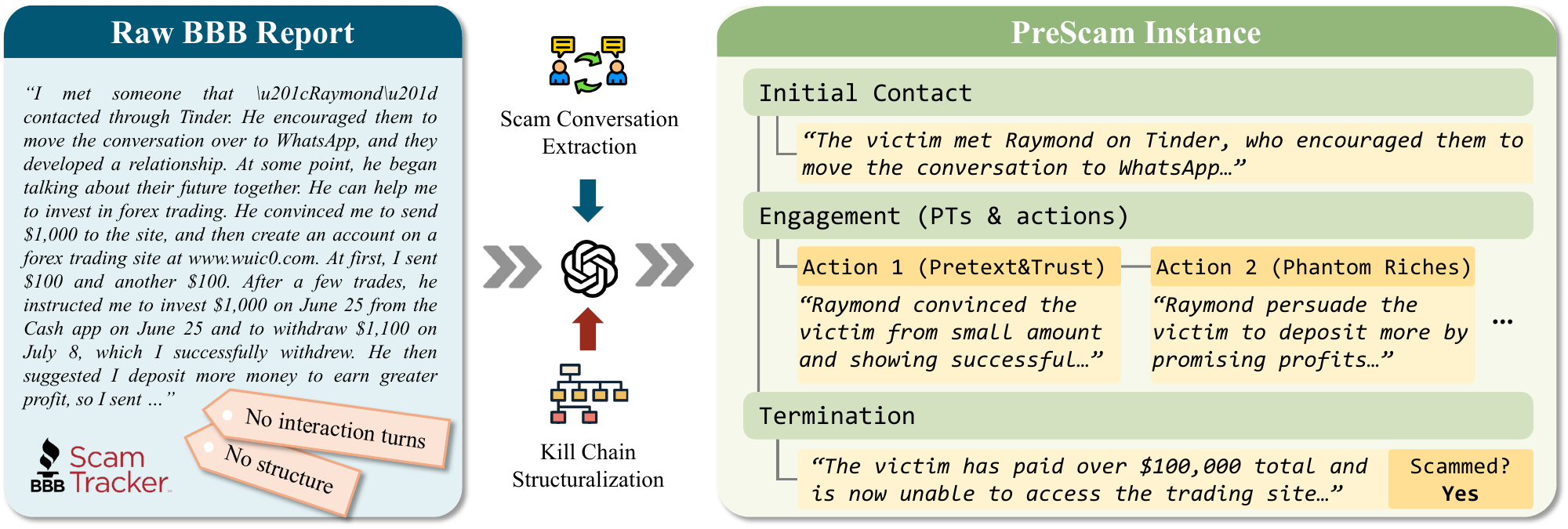}
    \caption{From Raw BBB Report to \bench Instance: an example of scam conversation extraction and kill chain structuralization. The left panel shows an unstructured user-submitted BBB report without explicit interaction turns, while the right panel shows the resulting \bench instance with Initial Contact, Engagement PT actions, and Termination.}
    \label{fig:kill chain}
\end{figure}

\section{Related Work}

\subsection{Scam Detection and Behavioral Analysis}

Current computer science research on online scams generally falls into two main categories: automated detection and behavioral analysis.

\paragraph{Detection.}
Automated detection of online scams has been studied across multiple modalities. Visual and brand-impersonation signals have driven a line of phishing website detectors, evolving from CNN-based logo matching~\citep{lin2021phishpedia} to knowledge-graph- ~\citep{zhao2023self,ju2023graphpatcher,qian2022co,ju2022grape,zhao2021multi} and LLM-augmented reference-based systems~\citep{liu2023knowledge,li2024knowphish,cao2025phishagent}. Scams propagated through SMS and telephony have also received attention, with works characterizing smishing infrastructure~\citep{nahapetyan2024sms} and classifying robocall content at scale~\citep{prasad2023diving}. In the cryptocurrency domain, detection efforts span Ponzi schemes~\citep{chen2018detecting}, pump-and-dump operations~\citep{xu2019anatomy}, and transaction-based phishing~\citep{TxPhishScope}. More recently, the threat of LLM-generated phishing has prompted both attack studies and LLM-based defenses~\citep{roy2024chatbots,koide2024chatspamdetector}.

\paragraph{Behavioral.}
Behavioral research has examined scams from both victim and scammer perspectives. On the victim side, studies investigate susceptibility factors~\citep{hanoch2021scams,ye2009intelligent} and how persuasion principles such as authority and scarcity are exploited~\citep{van2019cognitive}. On the scammer side,~\citet{herley2012nigerian} offers a game-theoretic account of self-selection in advance-fee fraud, while~\citep{miramirkhani2016dial} documents social engineering scripts through direct scammer conversation. Structured manipulation lifecycles have been characterized for romance scams and the emerging pig-butchering variant~\citep{acharya2024explorative}.

However, these lines of work predominantly target static artifacts such as webpages and messages, or rely on post-hoc victim reports. They provide limited support for modeling scams that unfold dynamically through multi-turn interactions.

\subsection{Scam Conversation Datasets}

Recognizing the scarcity of research on conversational scams, recent studies have sought to bridge this gap by contributing scam conversation datasets. Some approaches rely on LLMs to synthesize these interactions~\citep{eder2025honeypot,yang2025fraud,ma2025teleantifraud,kumarage2025personalized}. However, synthesized data often diverges from real-world scenarios and struggles to capture the rapid evolution and structural complexity of actual scams~\citep{ma-etal-2025-psyscam,chen2025sok}. Alternative efforts involve active engagement, deploying personas or honeypots to interact directly with scammers and collect real-world conversation data~\citep{perkins2021honeypots,spokoyny2025victim,acharya2024explorative}. In contrast, our work contributes both a new dataset of real-world scam conversations and a benchmark for evaluating whether models can track and forecast scam progression from partial context.

\section{Preliminary}

\subsection{From Kill Chains to Scam Progression}

Existing cyber attack models, such as the Cyber Kill Chain~\citep{cyberkillchain}, formalize attacks into distinct sequential phases and associate each phase with specific attacker tactics and techniques~\citep{attck,barnum2012standardizingstix}. Similarly, social engineering kill chains~\citep{montanez2022cyber,longtchi2024internet} model human-centric attacks as staged psychological manipulation, while recent work~\citep{ma-etal-2025-psyscam} operationalizes this perspective through explicit Psychological Techniques (PTs). These frameworks suggest that scam conversations should not be viewed as isolated messages, but as structured processes that evolve over time through changing tactics.

This perspective is especially important for conversational scams. Scam conversations are not random collections of messages: the scammer first establishes contact, then gradually builds trust and manipulates the victim through sustained interaction, and eventually attempts to extract money, sensitive information, or compliance. Different stages serve different objectives, and therefore involve different behavioral tactics and psychological techniques. If we treat an entire conversation as unstructured free text, we lose the temporal progression of the attack, the transition between stages, and the changing role of psychological manipulation over time.

\subsection{Scam Kill Chain Representation}

To capture this structure, we introduce the \textit{Scam Kill Chain}, a representation that maps the temporal phases of a scam conversation to actions driven by psychological techniques, which we refer to as \textit{PT actions}. The chain consists of three primary phases: \textit{Initial Contact}, \textit{Engagement}, and \textit{Termination}. Each phase is realized through one or more PT actions, where the scammer operationalizes a specific psychological technique to achieve a phase-specific objective.

This representation makes the temporal progression of a scam explicit while preserving the underlying psychological techniques exploited at each stage. Figure~\ref{fig:kill chain} provides a concrete example of this representation, and Appendix~\ref{apx:taxo_of_pt} lists the PT taxonomy that we extend from recent work~\citep{ma-etal-2025-psyscam}. Appendix~\ref{sec:appendix_killchain_examples} provides additional examples across other scam types, such as pig-butchering and employment scams.

A \textbf{scam conversation} $\mathcal{C} = (u_1, u_2, \ldots, u_T)$ is a multi-turn dialogue between a scammer and a victim. Let $\mathcal{P} = \{p_1, p_2, \ldots, p_K\}$ denote the set of $K$ psychological techniques drawn from the predefined taxonomy. A \textbf{PT action} is a tuple $a_t = (p_t, u_t)$, where $p_t \in \mathcal{P}$ is the exploited PT and $u_t$ is the scammer utterance that operationalizes it. A \textbf{scam kill chain} is a temporally ordered triple of phases $\mathcal{K} = \langle \phi_{\text{IC}} \rightarrow \phi_{\text{EG}} \rightarrow \phi_{\text{TM}} \rangle$, representing \textit{Initial Contact}, \textit{Engagement}, and \textit{Termination}, respectively. Each phase $\phi = [a_i, \ldots, a_j]$ is a contiguous sequence of PT actions.

The scam kill chain structurally formalizes the conversation by inducing a temporal partition over $\mathcal{C}$ such that the full sequence is the concatenation of its disjoint phases:
$$ \mathcal{C} = \phi_{\text{IC}} \oplus \phi_{\text{EG}} \oplus \phi_{\text{TM}} $$
Concurrently, this framework maps each utterance $u_t$ to its corresponding underlying psychological technique $p_t$ to yield the structured action sequence.



\section{\bench Dataset}

\subsection{Overview}

We construct \bench entirely from real-world scam reports collected from BBB Scam Tracker between February 2024 and November 2025. Starting from 177,989 raw reports, we first identify 25,402 candidate multi-turn scam conversations using an LLM-based extraction procedure. We then remove cases with fewer than two engagement rounds, retaining 13,007 conversations, and further clean null or malformed entries, yielding a final dataset of 11,573 structured scam instances.

Each instance in \bench includes a scam category label, the original victim narrative, and a structured three-stage representation consisting of \textit{Initial Contact}, \textit{Engagement}, and \textit{Termination}. The engagement stage is further decomposed into multi-round scammer and victim actions with supporting verbatim spans and PT annotations, and each instance also includes the binary \textit{scammed} label and \textit{scammed\_reason}. This design preserves the evidential content of the original reports while transforming them into a structured representation suitable for stage-level and tactic-level analysis.

The final dataset covers 20 scam categories and exhibits a pronounced long-tailed distribution. The structured scam conversations are typically short and multi-round, whereas the original victim narratives are substantially longer and noisier. More detailed dataset schema descriptions, static analyses, and category-level insights, including category proportions, interaction-length statistics, and PT usage patterns, are provided in Appendix~\ref{apx:dataset_insights}. Details of the construction process are presented in the next subsection.



\subsection{Construction Pipeline}

\paragraph{Step 1: Seed Dataset Collection.}
We begin with publicly available real-world scam reports from BBB Scam Tracker~\citep{bbbscamtracker}. We collect all reports published between February 2024 and November 2025, resulting in an initial corpus of 177,989 raw reports.

\paragraph{Step 2: Scam Conversation Extraction.}
Raw user reports present substantial challenges for automated analysis due to their noisy and highly unstructured nature. Victims often submit emotional free-form narratives, incomplete descriptions, or single-turn accounts rather than genuine  scam conversations (see Appendix~\ref{apx:noisy_examples} for examples). Our preliminary analysis shows that a large portion of raw reports do not contain usable multi-turn conversations, making manual annotation prohibitively slow and expensive. Moreover, rule-based filtering is ineffective, as scam conversations do not exhibit reliable length-based or string-pattern regularities.

To address this issue, we employ \textsc{GPT-4o-mini} as an LLM-based extractor to filter out noisy reports and identify cases containing scammer and victim multi-turn conversations. This step yields 25,402 candidate conversations. We then remove cases with fewer than two engagement rounds, retaining 13,007 samples for downstream structuralization. The extraction prompt is provided in Appendix~\ref{sec:appendix_extraction_prompt}.

\begin{figure}
    \centering
    \includegraphics[width=0.99\linewidth]{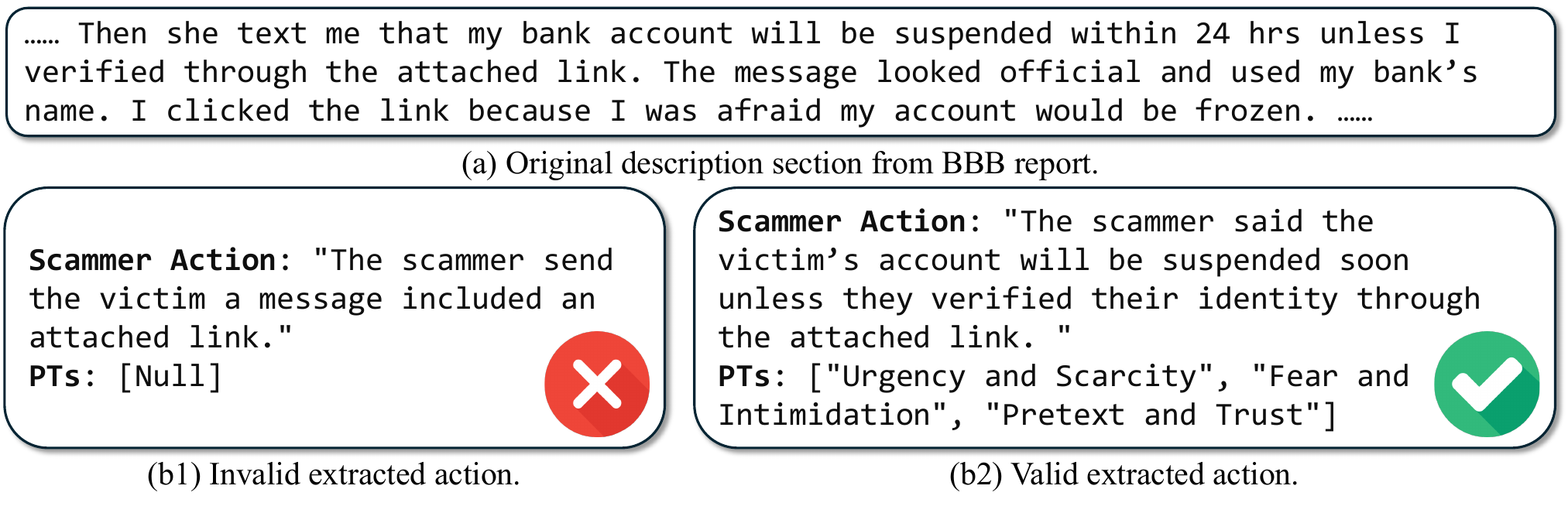}
    \caption{Comparison between invalid and valid extracted actions. An extracted action is considered valid only if it is grounded in the source report and paired with at least one PT label.}
    \label{fig:valid_invalid_action_comparison}
\end{figure}

\paragraph{Step 3: Kill Chain Structuralization.}
After extracting the core conversations, we map them onto our scam kill chain framework. Specifically, we use \textsc{MiniMax-2.5} to organize each case into three stages (i.e., Initial Contact, Engagement, and Termination) and identify the PT actions employed throughout the conversation. For the engagement stage, the model decomposes the conversation into fine-grained scammer and victim actions, aligns them with verbatim evidence spans, and assigns PT labels. Importantly, we define an extracted scammer action as \textit{valid} only when it is grounded in the source report and paired with at least one PT label; action spans with null PT annotations are treated as invalid and removed. Figure~\ref{fig:valid_invalid_action_comparison} illustrates this distinction with a concrete example. We further conduct post-processing to remove null or malformed entries introduced during generation, resulting in a final dataset of 11,573 structured instances. The structuralization prompt and formatting guidelines are provided in Appendix~\ref{sec:appendix_structuralization_prompt}.

\subsection{Quality Control}

To improve the reliability of the LLM-generated structures and mitigate potential hallucinations, we introduce a secondary LLM as a self-reflection agent to verify the structuralization outputs against a strict verification checklist including multiple perspectives.

We further conduct an independent human quality-control study on $n=200$ randomly sampled generated data. Three PhD-level reviewers evaluate each sample using an 8-point rubric. Averaged across reviewers, the mean score increases from $7.10$ to $7.37$ (out of 8), yielding a $+0.27$ improvement from the self-reflection stage. Reviewer-wise results, including win/tie statistics, per-dimension breakdowns, and the full evaluation criteria, are reported in Appendix~\ref{apx:human_review}.





\section{Benchmark Task Formulation}


We design two core tasks on the \bench dataset: \textit{Real-time Termination Prediction} and \textit{Scammer Action Prediction}. Both tasks operate over kill-chain-structured conversations.
The two tasks probe complementary aspects of scam progression modeling. Real-time termination prediction evaluates whether a model can track escalating scam risk from partial context, while scammer action prediction evaluates whether it can recover the future action--technique structure that drives that progression.

\subsection{Real-time Termination Prediction}

\paragraph{Motivation.}
In dynamic, interactive environments, such as live banking platforms or messaging applications~\citep{scamshield}, systems cannot always wait for the \textit{Termination} phase (e.g., the scammer explicitly asks for payment) before flagging a conversation as dangerous. Therefore, the goal is to predict whether the conversation is approaching the critical termination phase early enough to support timely intervention.


\paragraph{Task Definition.}
We frame real-time termination prediction as a \textit{continuous risk estimation} task. Given the conversational history up to turn $t$, denoted as $\mathcal{C}_{\leq t} = (u_1, \dots, u_t)$, the objective is to output a continuous risk score $\hat{p}_t \in [0, 1]$ indicating the proximity to the \textit{Termination} phase $\phi_{\text{TM}}$. A score of $\hat{p}_t \rightarrow 1$ indicates high confidence that the scammer's next turn will enter the termination phase (e.g., asking for payment). We evaluate this task under two inference settings: \textit{Direct LLM Prompting}, where an off-the-shelf LLM predicts whether the next turn enters the \textit{Termination} phase and assigns a risk score $\hat{p}_t = \text{LLM}(\mathcal{C}_{\leq t})$; and \textit{Supervised Sequence Classification}, where a neural network $f_{\theta}$ maps the conversation history directly to a continuous risk score $\hat{p}_t = f_{\theta}(\mathcal{C}_{\leq t})$.

\paragraph{Metrics.}
To evaluate the model's ability to maintain an accurate and timely rolling risk assessment, we utilize the following metrics:
\begin{itemize}[noitemsep, topsep=1pt, partopsep=1pt, listparindent=\parindent, leftmargin=*]
    \item \textbf{Area Under the Risk Curve (AUC):} We compute the aggregate AUC of the predicted trajectory $(\hat{p}_1, \dots, \hat{p}_{T_{\text{TM}}})$ to evaluate the overall monotonicity and confidence accumulation as the progresses.
    \item \textbf{Area Under the Precision-Recall Curve (AUPR):} Because the positive class (turns near the termination phase) is sparse - comprising only the final $k$ turns of each conversation - ROC-AUC can be overly optimistic under this class imbalance. We therefore also report AUPR (equivalent to average precision), which summarises the precision-recall trade-off across all thresholds and is more sensitive to performance on the minority positive class:
    $$ \text{AUPR} = \int_0^1 \mathrm{Precision}\bigl(\mathrm{Recall}^{-1}(r)\bigr)\,dr $$
    A higher AUPR indicates better identification of high-risk turns with fewer false alarms.
    \item \textbf{Alert Time (AT@FPR$_\alpha$):} To measure practical early-warning utility while ensuring comparability across methods, we select a method-specific threshold $\tau$ such that the false positive rate on the test set equals a fixed target $\alpha$ (we use $\alpha = 10\%$). Alert Time is then defined as the number of turns prior to the termination phase at which the risk score first breaches this threshold:
    $$ \text{AT@FPR}_\alpha = T_{\text{TM}} - \min\bigl\{t \mid \hat{p}_t \geq \tau_\alpha\bigr\}, \quad \tau_\alpha = \inf\bigl\{\tau : \mathrm{FPR}(\tau) \leq \alpha\bigr\}. $$
    A larger AT@FPR$_\alpha$ indicates an earlier and more actionable alert. Since $\tau_\alpha$ is calibrated to the same false positive rate for every method, AT@FPR$_\alpha$ is directly comparable across methods.

\end{itemize}


\subsection{Scammer Action Prediction}




\paragraph{Motivation.} During the Engagement phase, anticipating a scammer's next move is critical for prevention. If a system can forecast the scammer's likely subsequent action from the observed conversation history, it can provide an early warning before the interaction escalates to termination.


\paragraph{Task Definition.}
We frame this task as a sequential forecasting problem. Given the conversational context up to turn $t$, denoted as $\mathcal{C}_{\leq t} = (u_1, \dots, u_t)$, the objective is to predict the scammer's subsequent actions in the remaining conversation. We do not use a globally fixed boundary turn: as illustrated in Figure~\ref{fig:kill chain}, the scam intent is often hard to infer from the beginning of an interaction, since the \textit{Initial Contact} phase, and sometimes even the first \textit{Engagement} round, may still appear ambiguous. Different cases also reveal scam intent at different speeds, so prediction should start from a more reasonable boundary rather than from a fixed early turn. We therefore instantiate $t$ dynamically by first using \textsc{GPT-4o-mini} to label whether each observed prefix already appears likely to be a scam, and then training a separate \textsc{RoBERTa} classifier on these labels to select $t$ with a conservative threshold of 0.9, thereby improving generalization beyond the annotating LLM. 

We evaluate this task in a zero-shot setting under two inference conditions: \textit{Unlimited}, where the LLM is given the observed conversation context and generates future scammer actions freely, i.e., $\hat{A}_{> t} = \text{LLM}(\mathcal{C}_{\leq t})$; and \textit{Limited}, where the LLM is given the observed conversation context together with the exact number of remaining scammer actions and generates that many future actions, i.e., $\hat{A}_{> t} = \text{LLM}(\mathcal{C}_{\leq t},\; n_{\text{remain}})$, where $n_{\text{remain}}$ denotes the number of future scammer actions in the gold continuation after turn $t$.
Here, $\hat{A}_{> t}$ denotes the predicted sequence of future scammer actions. Additional details on this dynamic boundary construction are provided in Appendix~\ref{sec:appendix_action_prediction_details}.

\paragraph{Metrics.}
To evaluate predicted actions, we report two task-specific hit-rate metrics together with two standard text similarity metrics. Let the gold next PT action for evaluation instance \(i\) be \(a_i=(p_i,u_i)\), where \(u_i\) is the reference scammer utterance, and let the predicted next PT action be \(\hat{a}_i=(\hat{p}_i,\hat{u}_i)\). Because a generated next move may partially recover the intended action content or psychological techniques without exactly matching the reference wording, we first use an \textbf{LLM-as-a-Judge} framework with \textsc{GPT-4o-mini} to extract structured labels from the generated and gold next-action texts, and then compute hit-rate metrics over these extracted sets. Human review on 100 sampled cases (199 actions) shows strong agreement with the judge, reaching 92.0\% action-level agreement and Cohen's $\kappa=0.774$ (Appendix~\ref{apx:human_review}).

Specifically, let \(\Psi(\cdot)\) denote the set of atomic action elements extracted by the LLM judge from a next-action text, and let \(\Gamma(\cdot)\) denote the set of PT labels extracted by the same judge.

\begin{itemize}[noitemsep, topsep=1pt, partopsep=1pt, listparindent=\parindent, leftmargin=*]

\item \textbf{Action HitRate:}
We measure how much of the ground-truth action content is recovered by the prediction using the final score:
$$ \text{Action HitRate} = \frac{1}{N}\sum_{i=1}^{N} \frac{|\Psi(u_i)\cap \Psi(\hat{u}_i)|}{|\Psi(u_i)|} $$

\item \textbf{PT HitRate:}
Similarly, we measure how many ground-truth psychological techniques are recovered by the prediction using the final score:
$$ \text{PT HitRate} = \frac{1}{N}\sum_{i=1}^{N} \frac{|\Gamma(u_i)\cap \Gamma(\hat{u}_i)|}{|\Gamma(u_i)|} $$

\item \textbf{Auxiliary Text Metrics:}
In addition to the two judge-based hit-rate metrics above, we report BERTScore and ROUGE-L between the predicted utterance \(\hat{u}_i\) and the reference utterance \(u_i\) to measure semantic similarity and lexical overlap, respectively. We also report Precision to measure how much of the predicted action content is supported by the gold action set, i.e., the fraction of extracted predicted action elements that overlap with \(\Psi(u_i)\).

\end{itemize}

\section{Evaluation}

\subsection{Experiment Setups}

\paragraph{Real-time Termination Prediction. }We employ multiple baseline models categorized into three types for comparison: classical machine learning approaches (a position-only classifier using round index with logistic regression, and TF-IDF encoding~\citep{tfidf} + Logistic Regression) trained using an 80\%–20\% train–test split; neural models (MLP with TF-IDF features, MLP with embeddings, LSTM~\citep{Hochreiter1997LSTM}, hierarchical encoder, Transformer~\citep{vaswani2017attention}, and BERT-base-uncased~\citep{devlin2019bert}) fine-tuned under the same 80\%–20\% split; and LLMs (\textsc{GPT-4o-mini}~\citep{openai2024gpt4omini}, \textsc{DeepSeek-V3}~\citep{liu2025deepseek}, \textsc{Qwen3-235B-A22}~\citep{yang2025qwen3}, and \textsc{Grok-4.1-fast}~\citep{x2025grok41fast}) evaluated in a zero-shot setting, where models generate a risk score between 0 and 1 based on the prompt. To address class imbalance in supervised models, we apply techniques such as weighted loss functions and data resampling. More details are shown in Appendix~\ref{sec:appendix_rt_prediction_details}.

\paragraph{Scammer Action Prediction. }Given the conversation up to turn $t$, models are tasked with generating the scammer's subsequent actions for the remaining conversation. We evaluate multiple LLMs in a zero-shot setting: \textsc{GPT-4o-mini}, \textsc{GPT-5}, \textsc{Claude-Sonnet-4.5}, \textsc{DeepSeek-V3.2}, and \textsc{Llama-3.3-70B-Instruct}. Each model receives the observed conversation prefix up to $t$ and generates a structured sequence of predicted scammer actions. Action HitRate and PT HitRate are computed using an LLM-as-a-Judge with \textsc{GPT-4o-mini}. Detailed human-validation results for this judge and other evaluation details can be found in Appendix~\ref{apx:human_review} and Appendix~\ref{sec:appendix_action_prediction_details}.

\begin{wraptable}{r}{0.60\linewidth}
\centering
\resizebox{0.99\linewidth}{!}{
\begin{tabular}{lccc}
\toprule
\textbf{Methods} & \textbf{AUC (\%)$\uparrow$} & \textbf{AUPR (\%)$\uparrow$} & \textbf{AT@FPR$_{10\%}$$\uparrow$} \\
\midrule
\rowcolor{gray!10}
position\_only & 77.6 & 53.3 & 1.76 \\
TF-IDF & 81.1 {\small\textcolor{green!60!black}{+3.5}} & 61.6 {\small\textcolor{green!60!black}{+8.3}} & 1.58 {\small\textcolor{red!70!black}{-0.18}} \\
mlp\_tfidf & 79.6 {\small\textcolor{green!60!black}{+2.0}} & 59.0 {\small\textcolor{green!60!black}{+5.7}} & 1.60 {\small\textcolor{red!70!black}{-0.16}} \\
mlp\_embed & 81.6 {\small\textcolor{green!60!black}{+4.0}} & 61.9 {\small\textcolor{green!60!black}{+8.6}} & 1.57 {\small\textcolor{red!70!black}{-0.19}} \\
LSTM & 79.5 {\small\textcolor{green!60!black}{+1.9}} & 56.5 {\small\textcolor{green!60!black}{+3.2}} & 1.73 {\small\textcolor{red!70!black}{-0.03}} \\
hierarchical & 79.5 {\small\textcolor{green!60!black}{+1.9}} & 60.2 {\small\textcolor{green!60!black}{+6.9}} & 1.60 {\small\textcolor{red!70!black}{-0.16}} \\
transformer & 79.6 {\small\textcolor{green!60!black}{+2.0}} & 56.7 {\small\textcolor{green!60!black}{+3.4}} & 1.71 {\small\textcolor{red!70!black}{-0.05}} \\
Bert & \textbf{83.4} {\small\textcolor{green!60!black}{+5.8}} & \textbf{65.6} {\small\textcolor{green!60!black}{+12.3}} & 1.50 {\small\textcolor{red!70!black}{-0.26}} \\
\textsc{GPT-4o-mini} & 67.4 {\small\textcolor{red!70!black}{-10.2}} & 44.9 {\small\textcolor{red!70!black}{-8.4}} & 1.80 {\small\textcolor{green!60!black}{+0.04}} \\
\textsc{DeepSeek-V3.2} & 69.9 {\small\textcolor{red!70!black}{-7.7}} & 47.3 {\small\textcolor{red!70!black}{-6.0}} & 1.74 {\small\textcolor{red!70!black}{-0.02}} \\
\textsc{Qwen3-235B-A22} & 68.7 {\small\textcolor{red!70!black}{-8.9}} & 45.2 {\small\textcolor{red!70!black}{-8.1}} & 1.33 {\small\textcolor{red!70!black}{-0.43}} \\
\textsc{Grok-4.1-fast} & 63.3 {\small\textcolor{red!70!black}{-14.3}} & 41.9 {\small\textcolor{red!70!black}{-11.4}} & \textbf{2.03} {\small\textcolor{green!60!black}{+0.27}} \\
\bottomrule
\end{tabular}
}
\caption{Performance comparison of methods ordered from simple heuristics to large-scale language models.}
\label{tab:re_detect}
\end{wraptable}

\subsection{Main Results}

\paragraph{Real-time Termination Prediction}


Table~\ref{tab:re_detect} shows that real-time termination prediction is driven much more by task-specific progression modeling than by general-purpose LLM reasoning. Overall, supervised encoders substantially outperform zero-shot LLMs on the main ranking metrics, indicating that robust turn-level risk estimation requires learning sequential risk signals from data rather than relying on broad scam-related world knowledge alone. Specifically, first, most supervised text models outperform the position-only baseline on both AUC and AUPR, showing that conversational content provides strong information beyond simple turn position. Second, \textsc{BERT} achieves the best overall ranking performance, reaching 83.4 AUC and 65.6 AUPR, which suggests that pretrained bidirectional encoders are especially effective at aggregating local semantic cues from partial scam histories. Moreover, zero-shot frontier LLMs perform markedly worse on these two metrics, with all of them falling below the simple position-only baseline in both AUC and AUPR. An interesting finding, however, is that the ranking changes for alert timeliness: although \textsc{BERT} is strongest in overall discrimination, \textsc{Grok-4.1-fast} achieves the highest AT@FPR$_{10\%}$, implying that some LLMs are more willing to issue earlier warnings even when their overall risk calibration remains weak.

\paragraph{Scam Action Prediction}
\begin{table}[]
    \centering
    \resizebox{\linewidth}{!}{
    \begin{tabular}{clccccc}
    \toprule
        \textbf{Setting} & \textbf{Method} & \textbf{AHit$(\%)$} & \textbf{PTHit$(\%)$} & \textbf{BERTScore} &  \textbf{ROUGE-L} & \textbf{Precision} \\
        \midrule
        \multirow{5}{*}{Unlimited} & \textsc{GPT-5-chat} & 71.84 & 58.23 & \textbf{0.8603} & \textbf{0.1508} & \textbf{0.7323} \\
         & \textsc{GPT-4o-mini} & 65.15 & 57.55 & 0.8593 & 0.1446 & 0.7166 \\
         & \textsc{Claude-Sonnet-4.5} & \textbf{79.36} & 53.76 & 0.8526 & 0.1343 & 0.7052 \\
         & \textsc{DeepSeek-V3.2} & 75.80 & \textbf{60.43} & 0.8439 & 0.1134 & 0.5377 \\
         & \textsc{Llama-3.3-70B} & 74.80 & 55.59 & 0.8527 & 0.1379 & 0.6732 \\
         \midrule
         \multirow{5}{*}{Limited} & \textsc{GPT-5-chat} & 57.53 & \textbf{50.91} & 0.8680 & 0.1865 & 0.6673 \\
         & \textsc{GPT-4o-mini} & 40.45 & 45.63 & 0.8740 & \textbf{0.2075} & 0.5963 \\
         & \textsc{Claude-Sonnet-4.5} & \textbf{63.93} & 43.22 & 0.8661 & 0.1896 & \textbf{0.7345}\\
         & \textsc{DeepSeek-V3.2} & 63.57 & 49.67 & 0.8567 & 0.1588 & 0.7121 \\
         & \textsc{Llama-3.3-70B} & 52.52 & 46.43 & \textbf{0.8686} & 0.1976 & 0.6918 \\

         \bottomrule
    \end{tabular}
    }
    \caption{Scammer action prediction performance under the Unlimited and Limited settings. We report Action HitRate (AHit), PT HitRate (PTHit), BERTScore, ROUGE-L, and Precision for each model.}
    \label{tab:action pred}
\end{table}

\begin{figure}
    \centering
    \includegraphics[width=0.99\linewidth]{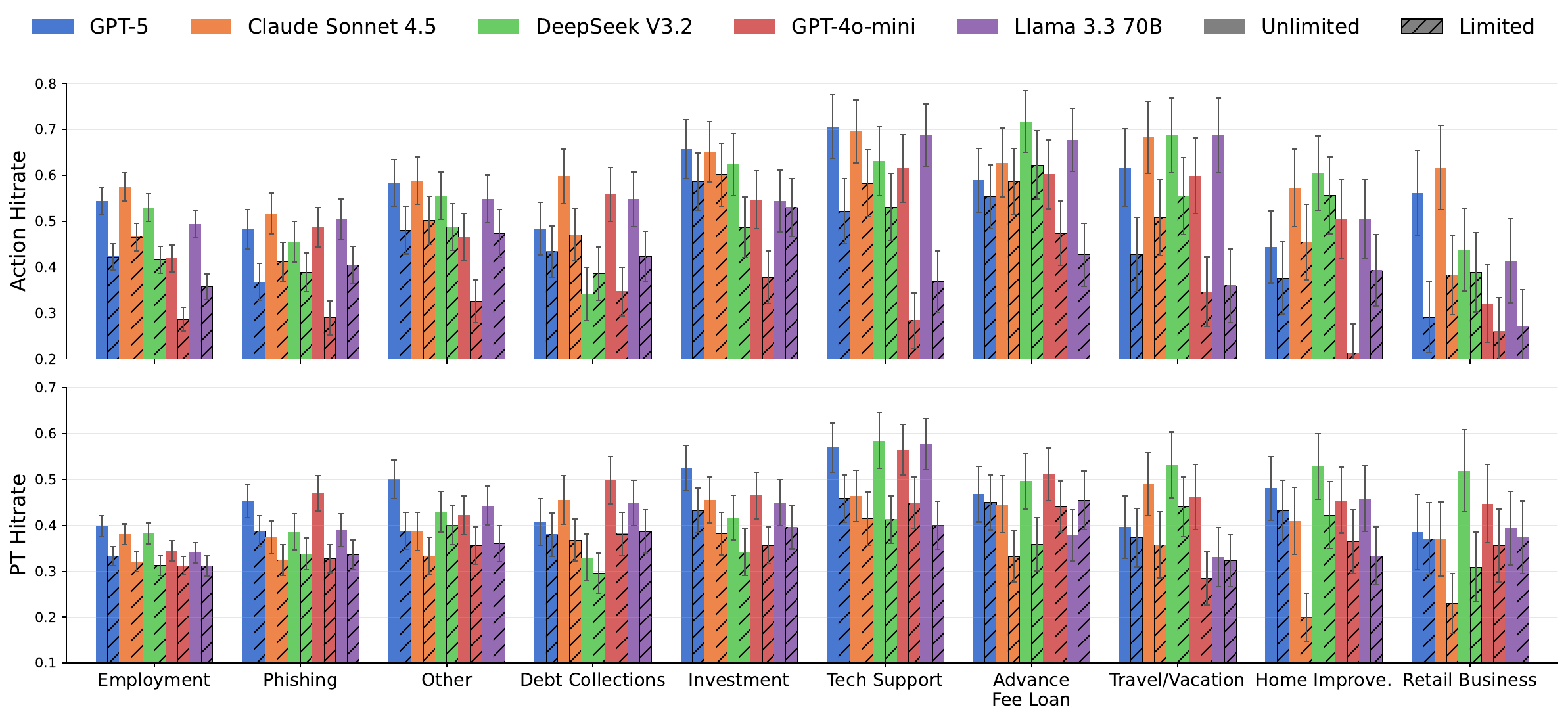}
    \caption{Per-scam-type breakdown of scammer action prediction performance. The top panel reports Action HitRate, and the bottom panel reports PT HitRate across scam categories. Colors indicate models, while solid and hatched bars denote the Unlimited and Limited settings, respectively.}
    \label{fig:by scam type}

\end{figure}

Table~\ref{tab:action pred} and Figure~\ref{fig:by scam type} show that scammer action prediction remains challenging even for frontier LLMs. Overall, the results indicate that the main bottleneck is not surface-form similarity, but correctly recovering the latent action--technique structure that drives scam progression. Specifically, first, under the \textit{Unlimited} setting, \textsc{Claude-Sonnet-4.5} achieves the highest overall Action HitRate (79.36), while \textsc{DeepSeek-V3.2} attains the best overall PT HitRate (60.43), again revealing a gap between recovering the concrete subsequent action and recovering its underlying psychological technique. Second, lexical overlap metrics such as ROUGE-L and BERTScore do not align with these structurally grounded measures: models in the \textit{Limited} setting often obtain higher BERTScore and ROUGE-L despite much lower AHit and PTHit. This pattern suggests that constraining the number of future actions mainly regularizes surface realization, making outputs look closer to the reference without necessarily recovering the correct action decomposition or technique sequence. 

Moreover, restricting the output space further hurts action recovery, and the magnitude of this drop varies across scam categories. The recurring gap between AHit and PTHit also suggests that models often capture the broad manipulative intent of a continuation before they recover the exact operational step that advances the scam. An interesting finding is that this decoupling also appears at the model level: under the \textit{Limited} setting on Tech Support, \textsc{GPT-4o-mini} achieves a near-best PT HitRate (44.87\%) despite the lowest AHit (28.38\%) among all models on that category, indicating that recognizing the underlying technique can be easier than forecasting the exact actions.

\paragraph{Overall Findings.}
Across both tasks, the same weakness emerges. Current models struggle not only to track how risk builds toward termination in real time, but also to recover the action and technique progression that drives scam escalation, suggesting that the core limitation lies in structural progression modeling rather than in recognizing isolated scam cues alone. Scam understanding appears easier at the level of local cues or broad manipulative intent than at the level of temporally coherent progression. Another is that stronger surface generation does not necessarily translate into better scam forecasting: models can produce plausible continuations while still missing when the scam is escalating and what concrete step comes next.




\section{Conclusion}

This work reframes conversational scam understanding as a progression modeling problem rather than a static text classification problem. By introducing \bench and its two complementary tasks, we make it possible to study not only whether a model recognizes scam-related cues, but also whether it can tell when a conversation is becoming dangerous and what the scammer is likely to do next. Our results show that these abilities remain far from solved: strong language models can often generate plausible continuations or capture parts of the underlying technique, yet they still struggle to track escalation and recover the action structure of real scam conversations. We hope \bench\ can help the community move toward progression-aware scam analysis, more rigorous evaluation of interactive risk understanding, and ultimately more timely intervention systems for real-world scam settings.

\section*{Ethics Statement}
This work uses scam reports that are publicly accessible through BBB Scam Tracker. We use these reports solely for academic research on scam understanding, scam progression modeling, and defensive evaluation. Our goal is to study whether models can better identify evolving scam risk and anticipate scammer behavior in realistic conversations, rather than to support real-world scam deployment or persuasive message generation.

Our benchmark includes structured representations of scammer actions and psychological techniques, and our action-prediction experiments require models to generate scammer-side continuations. These generations are produced only in a controlled offline research setting for evaluation purposes. We emphasize that \bench\ is intended to support defensive research, including risk forecasting, scam analysis, and the development of safer detection and intervention systems. Because the underlying reports describe harmful real-world events, the data and derived benchmark should not be repurposed to facilitate deceptive, manipulative, or otherwise harmful applications.

\bibliography{colm2026_conference}
\bibliographystyle{colm2026_conference}

\newpage
\appendix
\onecolumn

\label{sec:appendix}



\label{appendix:convacc}

\section{Taxonomy of PTs}
\label{apx:taxo_of_pt}
\begin{table*}[htbp]
\centering
\renewcommand{\arraystretch}{1.2}
\begin{tabularx}{\textwidth}{|l|X|}
\hline
\textbf{Principle} & \textbf{Description} \\
\hline
Authority & From Cialdini's 6 principles of persuasion: People tend to obey authorities and trust credible individuals. \\
\hline
Phantom Riches & Visceral triggers of desire that override rationality. \\
\hline
Fear and Intimidation & Leverages the fear response which overrides rational thought. \\
\hline
Liking & From Cialdini's 6 principles of persuasion: Preference for saying “yes” to requests from people they know and like. People are programmed to like others who like them back and who are similar to them. \\
\hline
Urgency and Scarcity & From Cialdini's 6 principles of persuasion: A sense of urgency and scarcity assigns more value to items. \\
\hline
Pretext and Trust & Scammers make up stories to add source credibility and gain victims' trust. \\
\hline
Evoking Social Norms & Exploiting socially desired rules or norms such as being kind, giving, helpful, or the tendency to feel obliged to repay favors from others (``I do something for you, you do something for me.''). \\
\hline
Consistency & From Cialdini's 6 principles of persuasion: Tendency to behave in a way consistent with past decisions and behaviors. \\
\hline
Social Proof & From Cialdini's 6 principles of persuasion: Tendency to reference the behavior of others, using majority behavior to guide actions. \\
\hline
\end{tabularx}
\caption{Taxonomy of PTs}
\end{table*}

\section{Additional Dataset Statistics and Insights}
\label{apx:dataset_insights}

\subsection{Construction Summary}

\bench is constructed entirely from real-world scam reports collected from BBB Scam Tracker between February 2024 and November 2025. Starting from 177,989 raw reports, we first identify 25,402 candidate multi-turn scam conversations using an LLM-based extraction procedure. We then remove cases with fewer than two engagement rounds, retaining 13,007 conversations, and further clean null or malformed entries, yielding a final dataset of 11,573 structured scam instances.

\subsection{Instance Structure}

Each instance in \bench contains the following seven components:
\begin{itemize}[noitemsep, topsep=2pt, partopsep=0pt, leftmargin=*]
    \item a scam category label;
    \item the original victim narrative;
    \item an \textit{Initial Contact} summary describing how the scam begins;
    \item an \textit{Engagement} sequence consisting of multi-round scammer and victim actions with supporting verbatim spans and PT annotations;
    \item a \textit{Termination} summary describing how the scam reaches its final decisive stage;
    \item a binary \textit{scammed} label indicating whether the scammer succeeded; and
    \item a \textit{scammed\_reason} field explaining the decision.
\end{itemize}
This design preserves the evidential content of the original reports while transforming them into a structured representation suitable for stage-level and tactic-level analysis.

\subsection{Static Statistics}

\bench spans 20 scam categories and exhibits a pronounced long-tailed distribution. Employment scams dominate the corpus (31.6\%), followed by Phishing (12.3\%), while the remaining categories each account for much smaller shares. The number of engagement rounds ranges from 3 to 15, with a mean of 3.87, and most conversations concentrate between 3 and 5 rounds. The original victim descriptions are substantially noisier and longer, with an average length of 274.7 words, a median of 208 words, and extreme outliers reaching 10,266 words.

\begin{figure}[t]
\centering
\includegraphics[width=0.72\linewidth]{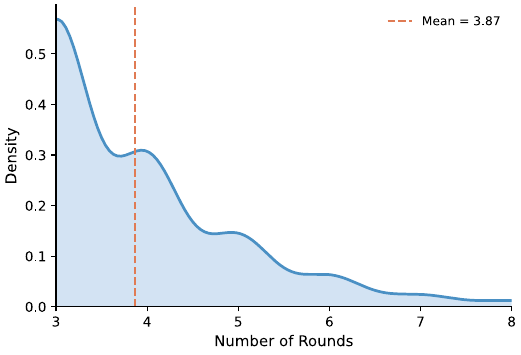}
\caption{Distribution of the number of engagement rounds in \bench.}
\label{fig:basic_stats}
\end{figure}

\begin{figure}[t]
\centering
\includegraphics[width=\linewidth]{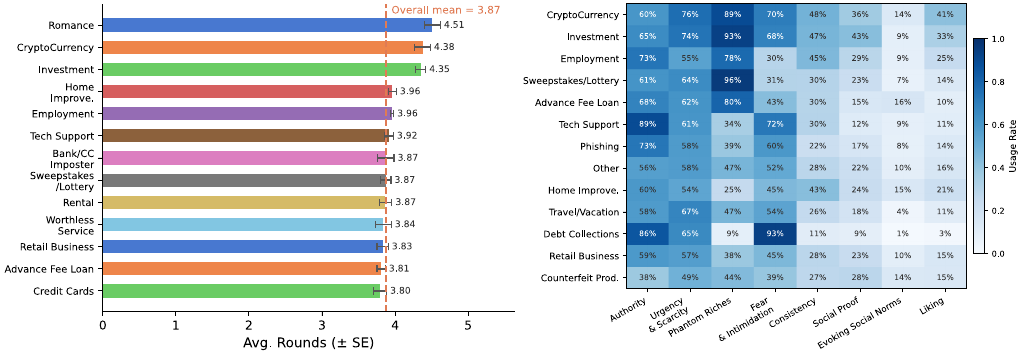}
\caption{Category-level conversation patterns in \bench. Left: average number of engagement rounds for each scam type. Right: PT usage rates across scam categories.}
\label{fig:pt_insight}
\end{figure}

Figure~\ref{fig:basic_stats} shows the overall distribution of engagement rounds in \bench. Figure~\ref{fig:pt_insight} further illustrates category-level variation in both interaction length and PT usage, highlighting substantial heterogeneity in real-world scam execution strategies.

\section{Examples of Noisy and Unstructured User Reports}
\label{apx:noisy_examples}

Raw user-submitted scam reports often lack standardized structure and vary 
substantially in format and content. The following examples illustrate 
common irregularities observed in real-world submissions.

\subsection{Single-Round Narrative Reports}

\textbf{Case 814017 (Investment, Reported Loss: \$1,048).}

\begin{leftbarquote}
“Told me that I had over \$80,000 in Bitcoin did I needed to exchange and I had to pay a fee of \$2,000 to access it and I didn't have the \$2,000 but I paid a little over \$1,000 as a down payment and then they never said anything else and it keeps sending me notices.”
\end{leftbarquote}

This report is written as a single retrospective paragraph and does not provide 
multi-turn conversation details. The description is informal and unstructured, 
requiring interpretation of the underlying event from free-form text.

\subsection{Category-Misaligned Reports}

\textbf{Case 848590 (Phishing, Reported Loss: \$60,000).}

\begin{leftbarquote}
“Human trafficking, by a human services agency of a child born 3-12-23 ... I filed complaint with cps in Bakersfield Roxanne.”
\end{leftbarquote}

Although labeled as phishing, the content does not clearly describe a scam scenario. 
Instead, the report appears to concern a complaint involving a public agency.

\subsection{Extremely Short and Underspecified Reports}

\textbf{Case 1079259 (Identity Theft, Reported Loss: \$4,500).}

\begin{leftbarquote}
“I was not paid for the work that I done.”
\end{leftbarquote}

This submission consists of a single short sentence without conversation context, 
timeline, or explicit description of a fraud mechanism. The narrative alone 
provides limited information for automated interpretation.

\section{Human Review Protocol and Results}
\label{apx:human_review}

\begin{figure}[t]
\centering

\begin{subfigure}[t]{0.46\linewidth}
    \centering
    \includegraphics[width=\linewidth]{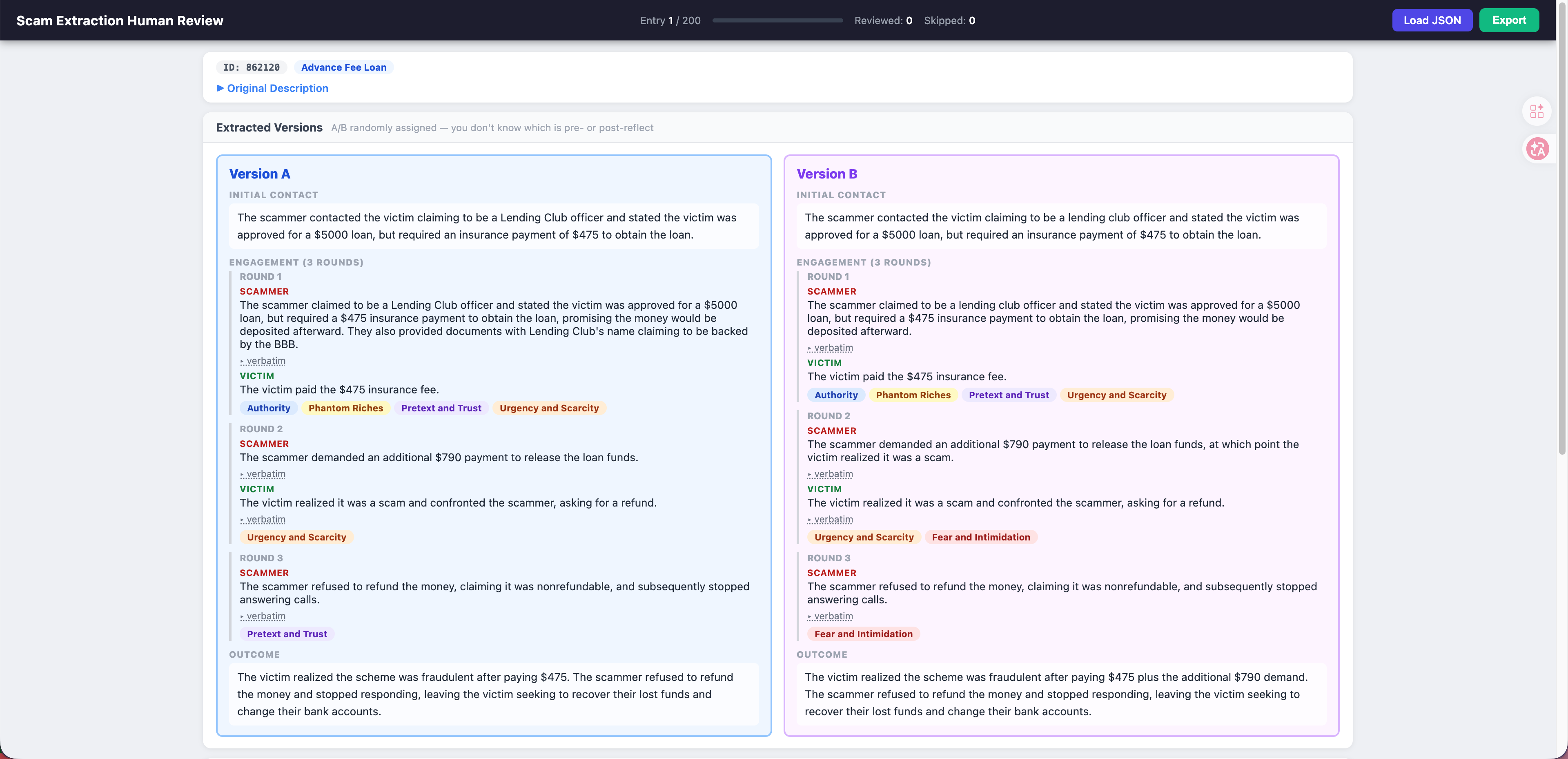}
    \caption{Comparison Page.}
    \label{fig:ui_a}
\end{subfigure}
\hfill
\begin{subfigure}[t]{0.52\linewidth}
    \centering
    \includegraphics[width=\linewidth]{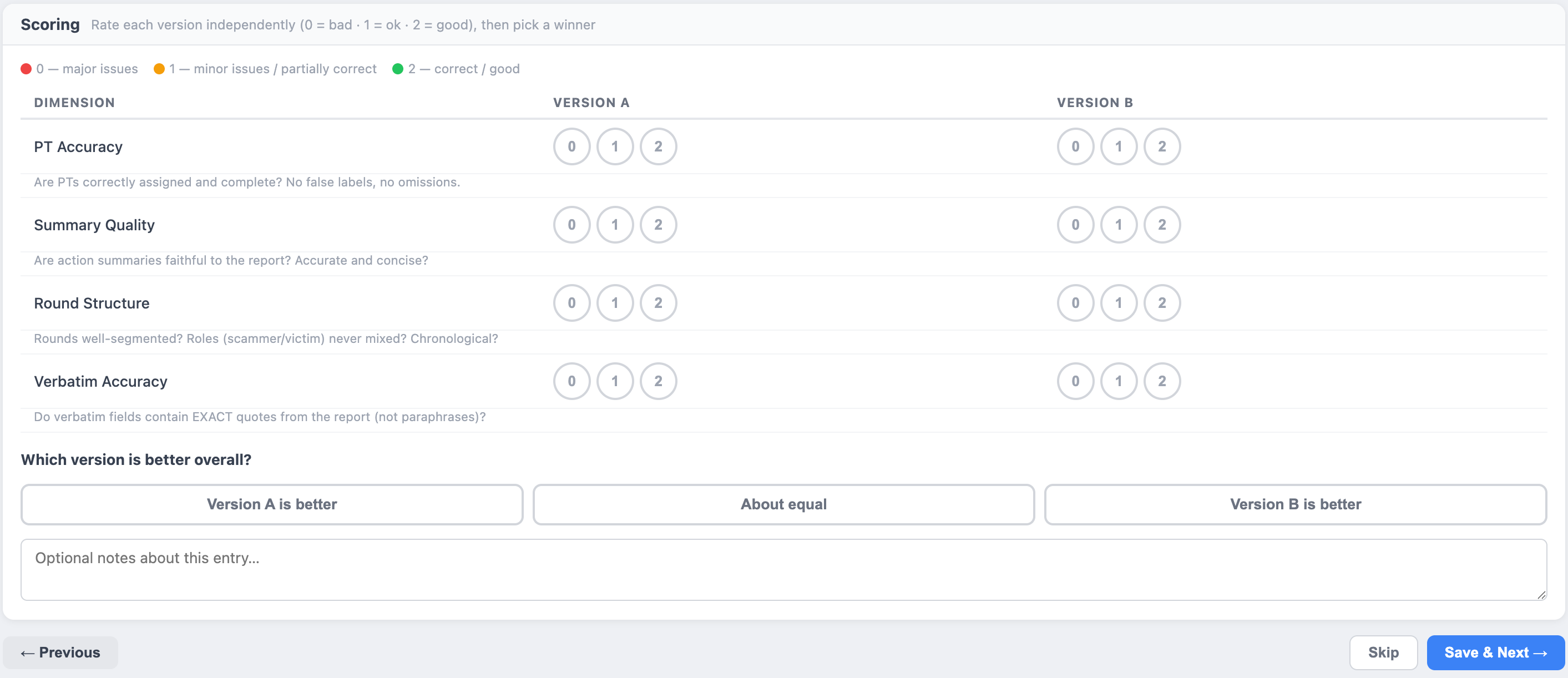}
    \caption{Scoring panel.}
    \label{fig:ui_b}
\end{subfigure}

\caption{Human evaluation interface used in the blind paired study. 
Pre/Post outputs are randomly assigned to A/B. Reviewers score four dimensions (0-2 scale) and provide a head-to-head winner judgment.}
\label{fig:human_interface}

\end{figure}

\subsection{Blind Paired Evaluation Protocol}

We conduct a blind paired human evaluation on $n=200$ randomly sampled generated kill chains. For each entry, the \textit{pre-reflection} and \textit{post-reflection} outputs are randomly assigned to anonymized labels (A/B). Reviewers are unaware of the mapping between A/B and the generation condition. The mapping is restored only after all annotations are finalized. The evaluation interface is shown in Figure~\ref{fig:human_interface}.

Three independent PhD-level reviewers evaluate all samples. Each reviewer:

\begin{itemize}
    \item Assigns a score on four dimensions (0-2 scale per dimension; total max = 8).
    \item Performs a head-to-head comparison: \textit{A better / Equal / B better}.
\end{itemize}

This design ensures unbiased evaluation and separates absolute scoring from pairwise preference judgment.

\subsection{Scoring Rubric}

Each dimension is scored on a 0-2 scale (maximum total score = 8).

\begin{table}[h]
\centering
\caption{Human evaluation dimensions and what they measure.}
\begin{tabular}{ll}
\toprule
Dimension & What It Measures \\
\midrule
PT Accuracy & Correctness and completeness of PT labels \\
Summary Quality & Faithfulness of the summary to the original report \\
Round Structure & Proper separation of rounds and role delineation \\
Verbatim Accuracy & Whether verbatim fields are quoted word-for-word \\
Winner (Head-to-head) & Direct comparison: A better / Equal / B better \\
\bottomrule
\end{tabular}
\end{table}

\subsection{Across-Reviewer Average}

Table~\ref{tab:human_avg} reports the average score across three reviewers.

\begin{table}[h]
\centering
\caption{Across-reviewer average scores (max 8).}
\label{tab:human_avg}
\begin{tabular}{lccc}
\toprule
 & Pre & Post & $\Delta$ \\
\midrule
Mean Score (max 8) & 7.10 & 7.37 & +0.27 \\
\bottomrule
\end{tabular}
\end{table}

\subsection{Reviewer-wise Detailed Results}
Table~\ref{tab:human_all_reviewers} reports the detailed results of all three human reviewers.
\begin{table*}[h]
\centering
\caption{Reviewer-wise detailed human evaluation results ($n=200$ each). Scores are on an 8-point scale (max 8).}
\label{tab:human_all_reviewers}
\small
\begin{tabular}{lccccccccc}
\toprule
& \multicolumn{3}{c}{Reviewer R1} 
& \multicolumn{3}{c}{Reviewer R2}
& \multicolumn{3}{c}{Reviewer R3} \\
\cmidrule(lr){2-4} \cmidrule(lr){5-7} \cmidrule(lr){8-10}
Metric 
& Pre & Post & $\Delta$
& Pre & Post & $\Delta$
& Pre & Post & $\Delta$ \\
\midrule

Mean Score (max 8)
& 6.59 & 7.01 & +0.420
& 7.33 & 7.48 & +0.150
& 7.365 & 7.605 & +0.240 \\

\midrule
PT Accuracy
& 1.135 & 1.380 & +0.245
& 1.675 & 1.735 & +0.060
& 1.545 & 1.685 & +0.140 \\

Summary Quality
& 1.805 & 1.905 & +0.100
& 1.950 & 1.980 & +0.030
& 1.940 & 1.990 & +0.050 \\

Round Structure
& 1.870 & 1.900 & +0.030
& 2.000 & 2.000 & +0.000
& 1.960 & 1.985 & +0.025 \\

Verbatim Accuracy
& 1.780 & 1.825 & +0.045
& 1.705 & 1.765 & +0.060
& 1.920 & 1.945 & +0.025 \\

\midrule
Post Wins (\%)
& \multicolumn{3}{c}{53.0\%}
& \multicolumn{3}{c}{37.0\%}
& \multicolumn{3}{c}{38.5\%} \\

Pre Wins (\%)
& \multicolumn{3}{c}{27.0\%}
& \multicolumn{3}{c}{31.0\%}
& \multicolumn{3}{c}{22.0\%} \\

Equal (\%)
& \multicolumn{3}{c}{20.0\%}
& \multicolumn{3}{c}{32.0\%}
& \multicolumn{3}{c}{39.5\%} \\

\bottomrule
\end{tabular}
\end{table*}

\subsection{Validation of the LLM Judge}

We use \textsc{GPT-4o-mini} as the LLM judge for action-level coverage evaluation. To assess its reliability, we manually review 100 sampled cases comprising 199 total gold actions and compare the judge's binary hit/miss decisions against a human reviewer. Table~\ref{tab:judge_human_agreement} summarizes the overall agreement statistics. At the action level, the judge reaches 92.0\% agreement with the reviewer and Cohen's $\kappa=0.774$, indicating substantial agreement. At the case level, 86 of the 100 sampled cases have exactly the same action-level judgments, and the mean absolute difference in per-case AHit is 0.081.

\begin{table}[t]
\centering
\caption{Overall agreement between the \textsc{GPT-4o-mini} judge and human review on 100 sampled cases (199 total actions).}
\label{tab:judge_human_agreement}
\begin{tabular}{lc}
\toprule
Metric & Value \\
\midrule
Total actions rated & 199 \\
Action-level agreement & 0.920 \\
Cohen's $\kappa$ & 0.774 \\
Precision & 0.935 \\
Recall & 0.960 \\
F1 & 0.948 \\
Full case agreement & 86/100 (86.0\%) \\
\bottomrule
\end{tabular}
\end{table}

Table~\ref{tab:judge_human_breakdown} further breaks down the agreement by the number of gold actions in a case and reports the action-level confusion counts. Agreement remains high across all buckets and is slightly stronger for cases with more gold actions.

\begin{table}[t]
\centering
\caption{Detailed agreement breakdown for the \textsc{GPT-4o-mini} judge.}
\label{tab:judge_human_breakdown}
\begin{tabular}{lccc}
\toprule
Bucket & $n$ & Agreement & $\kappa$ \\
\midrule
GT actions = 1 & 43 & 0.884 & 0.741 \\
GT actions = 2--3 & 103 & 0.922 & 0.761 \\
GT actions = 4+ & 53 & 0.943 & 0.822 \\
\midrule
\multicolumn{4}{l}{\textit{Action-level confusion matrix}} \\
\midrule
Original=Hit, Reviewer=Hit & \multicolumn{3}{c}{145} \\
Original=Hit, Reviewer=Miss & \multicolumn{3}{c}{6} \\
Original=Miss, Reviewer=Hit & \multicolumn{3}{c}{10} \\
Original=Miss, Reviewer=Miss & \multicolumn{3}{c}{38} \\
\bottomrule
\end{tabular}
\end{table}

\section{Extraction Prompt}
\label{sec:appendix_extraction_prompt}

We use \textsc{GPT-4o-mini} with the following prompt template in Step 2 to determine whether a raw BBB report contains usable multi-round interaction information.

\begin{verbatim}
You are an expert at analyzing scam victim reports. Your task is to
determine if the following report contains multi-round interaction
information.

Multi-round interaction means:
1. "multi-round dialogue": The report contains direct chat logs or
   conversation transcripts with back-and-forth messages between the
   victim and scammer.
2. "multi-round description": The report describes a sequence of
   multiple interactions over time, showing a process with multiple
   exchanges.

Single-round means: The report only describes a one-time event or a
single interaction without follow-up exchanges.

Report:
"""
{description}
"""

Analyze and respond in this exact format:
RESULT: [YES or NO]

Only output these two lines, nothing else.
\end{verbatim}

\section{Structuralization Prompt}
\label{sec:appendix_structuralization_prompt}

Step 3 uses \textsc{MiniMax-2.5} in a two-stage prompting pipeline. We first ask the model to convert each extracted multi-round report into a structured kill-chain instance, and then apply a reflection prompt to revise the output against a checklist. The placeholder \verb|{pt_definitions}| is instantiated with the PT taxonomy in Appendix~\ref{apx:taxo_of_pt}, and \verb|{scam_type}| is filled with the original BBB scam category. In the code, the final-stage field is named \verb|outcome|; in the paper, this field corresponds to the \textit{Termination} summary.

\paragraph{Initial Structuralization Prompt.}
\begin{verbatim}
You are an expert at analyzing scam victim reports. Your task is to
extract a structured breakdown of the scam from the victim's
description.

Structure to extract

1. initial_contact:
   A concise summary (1-2 sentences) of how the scammer first reached
   or lured the victim.

2. engagement:
   A list of interaction rounds where the scammer uses psychological
   techniques. Each round MUST have at least one PT. For each round,
   extract:
   - scammer_action: A concise summary (1-2 sentences) of what the
     scammer did or said in this round.
   - scammer_action_verbatim: The VERBATIM text from the report that
     corresponds to the scammer's action. Copy the exact words.
   - victim_action: A concise summary (1-2 sentences) of what the
     victim did or said in response (set to null if not present).
   - victim_action_verbatim: The VERBATIM text from the report that
     corresponds to the victim's action (set to null if not present).
     Copy the exact words.
   - PTs: List of psychological techniques the scammer used in this
     round (from the predefined list below). MUST contain at least one
     PT.

3. outcome:
   A concise summary (1-2 sentences) of the final result, such as the
   victim realizing it is a scam, losing money, or the scammer
   disappearing.

psychological techniques (PTs)
{pt_definitions}

Critical Rules
- For scammer_action and victim_action: use concise, factual summaries
  in 1-2 sentences.
- For scammer_action_verbatim and victim_action_verbatim: copy the 
  EXACT words from the report. Do NOT paraphrase.
- Every engagement round MUST have at least one PT. If a scammer action
  does not clearly use any persuasion technique, do NOT include it as a
  separate round; merge it into an adjacent round or place it in
  initial_contact or outcome as appropriate.
- An engagement round is defined by the scammer using a persuasion
  technique. Actions without psychological techniques are not
  engagement rounds.
- A single scammer_action can have multiple PTs.
- Keep the two roles strictly separated.
- Maintain chronological order.
- If a section is not present in the report, set it to null.

Output format
Return ONLY a JSON object:

{
  "initial_contact": "summary..." or null,
  "engagement": [
    {
      "scammer_action": "summary...",
      "scammer_action_verbatim": "exact words from report...",
      "victim_action": "summary..." or null,
      "victim_action_verbatim": "exact words from report..." or null,
      "PTs": ["PT_name_1", "PT_name_2"]
    }
  ],
  "outcome": "summary..." or null
}

Scam type: {scam_type}

Report:
"""
{description}
"""

Return the JSON object only.
\end{verbatim}

\paragraph{Reflection Prompt.}
After the initial extraction, we use a second prompt to refine the output through self-review:

\begin{verbatim}
You are an expert at analyzing scam victim reports. You previously
extracted a structured breakdown from a scam report. Now review your
extraction and fix any issues.

Original Report:
"""
{description}
"""

Your Previous Extraction:
{previous_result}

psychological techniques (PTs)
{pt_definitions}

Review Checklist
1. PT accuracy: Is each PT correctly assigned? Remove incorrect PTs and
   add missing ones.
2. PT completeness: Every engagement round MUST have at least one PT.
   If a round has no valid PT after review, merge it into an adjacent
   round or move it to initial_contact or outcome.
3. Round granularity: Should any rounds be split or merged?
4. Role separation: Is any victim behavior described in scammer_action,
   or vice versa?
5. Chronological order: Are the rounds in the correct time sequence?
6. Coverage: Are there persuasion actions from the report that were
   missed entirely?
7. initial_contact and outcome: Are they correct summaries? Should any
   content move between these fields and the engagement rounds?
8. Verbatim accuracy: Does each verbatim field contain EXACT text from
   the original report?

Output
Return the corrected JSON object only (same format as before). If no
changes are needed, return the same JSON.
\end{verbatim}

\section{Additional Scam Kill Chain Examples}
\label{sec:appendix_killchain_examples}

\section{Evaluation Details}
\label{sec:appendix_evaluation_details}

\subsection{Real-time Termination Prediction Details}
\label{sec:appendix_rt_prediction_details}

For real-time termination prediction, each evaluation instance consists of a partial scam conversation prefix together with a binary label indicating whether the scammer's next move enters the \textit{Termination} stage. Across all supervised methods, we use the same 80/20 train--test split at the \texttt{scam\_id} level, without a separate validation split. The input text is constructed by \texttt{format\_context()}, which concatenates the \textit{Initial Contact} summary with the observed round-level scammer and victim actions. For all PyTorch models, we apply gradient clipping with \texttt{max\_norm=1.0}, and model selection is based on the epoch with the lowest training loss rather than validation loss.

\paragraph{Classical Baselines.}
\begin{itemize}[noitemsep, topsep=2pt, partopsep=0pt, leftmargin=*]
    \item \textbf{Position-Only Baseline.} We fit a \texttt{StandardScaler} followed by logistic regression using only the scalar round index as input. The model outputs \texttt{predict\_proba} scores in $[0,1]$. No hyperparameter tuning is applied.
    \item \textbf{TF-IDF.} We use \texttt{TfidfVectorizer} with \texttt{max\_features=50{,}000}, \texttt{ngram\_range=(1,2)}, and \texttt{sublinear\_tf=True}, followed by logistic regression with \texttt{C=1.0} and \texttt{max\_iter=1000}. The classifier output is the predicted probability of entering termination at the next step.
\end{itemize}

\paragraph{Neural Models.}
Unless otherwise noted, the neural models are trained with AdamW (\texttt{lr=1e-3}, \texttt{weight\_decay=0.01}), cosine warmup scheduling, and \texttt{BCELoss}. The main differences lie in how conversation prefixes are encoded.

\begin{itemize}[noitemsep, topsep=2pt, partopsep=0pt, leftmargin=*]
    \item \textbf{MLP-TF-IDF.} Input features are TF-IDF vectors with \texttt{max\_features=10{,}000}, \texttt{ngram\_range=(1,2)}, and \texttt{sublinear\_tf=True}. The classifier is a two-hidden-layer MLP with hidden sizes 512 and 256, dropout 0.3, \texttt{batch\_size=64}, and 15 epochs.
    \item \textbf{MLP-Embed.} We use a word-level tokenizer with \texttt{vocab\_size=20{,}000} and \texttt{max\_len=256}. Tokens are mapped to 128-dimensional embeddings, mean-pooled over non-padding positions, and passed through the same two-hidden-layer MLP head used above. Training uses \texttt{batch\_size=64} for 15 epochs.
    \item \textbf{BiLSTM.} Using the same word-level tokenization, we encode each prefix with a two-layer bidirectional LSTM with hidden size 256 per direction, followed by a linear sigmoid classifier. Training uses \texttt{batch\_size=128} for 25 epochs.
    \item \textbf{Transformer (from scratch).} We again use the same word-level tokenizer and a learned token-plus-position embedding layer, followed by a two-layer Transformer encoder with \texttt{d\_model=128}, 4 attention heads, feed-forward dimension 512, and dropout 0.1. The encoded sequence is mean-pooled before binary classification. Training uses \texttt{batch\_size=128} for 25 epochs.
    \item \textbf{Hierarchical Encoder.} Each round is first encoded independently using the frozen sentence-transformer \texttt{all-MiniLM-L6-v2}; the \textit{Initial Contact} field is treated as round 0. The resulting round embeddings are then processed by a two-layer round-level Transformer encoder (\texttt{d\_model=384}, 4 heads, feed-forward dimension 256, dropout 0.1), followed by mean pooling over rounds and binary classification. Training uses \texttt{batch\_size=32} for 15 epochs.
    \item \textbf{BERT / RoBERTa.} We fine-tune pretrained encoders initialized from \texttt{bert-base-uncased} and \texttt{roberta-base} using the [CLS] representation with a dropout-plus-linear classification head. Inputs are tokenized with \texttt{max\_length=512}, truncation, and padding. Training uses AdamW with \texttt{lr=2e-5}, \texttt{weight\_decay=0.01}, cosine warmup, \texttt{batch\_size=16}, and 5 epochs.
\end{itemize}

\paragraph{Zero-shot LLMs.}
For zero-shot LLM evaluation, we provide the observed conversation history and ask the model to estimate the probability that the next move will enter the \textit{Termination} stage, returning a JSON object of the form \texttt{\{"risk\_score": <0-1>\}}. We use deterministic decoding (\texttt{temperature=0}) and evaluate these models directly on the test set without any task-specific training.

\subsection{Scammer Action Prediction Details}
\label{sec:appendix_action_prediction_details}

\paragraph{Dynamic Boundary Selection.}
We do not use a globally fixed boundary turn for scammer action prediction. As illustrated by Figure~\ref{fig:kill chain}, the \textit{Initial Contact} stage often does not make the scam intent explicit, and even the first \textit{Engagement} round may still be ambiguous. Moreover, scam conversations do not progress at a uniform speed: some cases reveal strong scam intent very early, while others unfold more gradually over several rounds. A fixed boundary would therefore create an uneven evaluation setup across cases.

To address this issue, we first use \textsc{GPT-4o-mini} to provide turn-level labels indicating whether the conversation already appears likely to be a scam at a given prefix. To improve generalization beyond the annotating LLM, we then train a RoBERTa classifier on these turn-level labels and use the resulting model to instantiate the boundary turn $t$ for downstream scammer action prediction. On the held-out test set at epoch 10, the classifier achieves 0.884 accuracy, 0.934 F1, and 0.881 ROC-AUC.

Table~\ref{tab:appendix_threshold_sweep} shows the threshold sweep used to convert RoBERTa scores into binary scam-likelihood decisions. As expected, higher thresholds improve precision but reduce recall and the fraction of prefixes marked as already scam-indicative. We use a threshold of 0.9 in the main experiments in order to define $t$ conservatively, i.e., only after the prefix has become highly likely to reflect an ongoing scam.


\begin{table}[t]
\centering
\caption{Threshold sweep for the RoBERTa-based boundary detector. \texttt{Triggered\%} denotes the fraction of prefixes classified as already likely to be scams.}
\label{tab:appendix_threshold_sweep}
\begin{tabular}{lcccc}
\toprule
Threshold & Precision & Recall & F1 & Triggered\% \\
\midrule
0.5 & 0.905 & 0.964 & 0.934 & 90.6\% \\
0.6 & 0.911 & 0.954 & 0.932 & 89.1\% \\
0.7 & 0.917 & 0.934 & 0.925 & 86.7\% \\
0.8 & 0.929 & 0.908 & 0.918 & 83.2\% \\
0.9 & 0.944 & 0.849 & 0.894 & 76.5\% \\
\bottomrule
\end{tabular}
\end{table}

Once the boundary turn has been instantiated, we use $t$ to denote the split between the observed prefix and the unobserved future continuation. For a structured scam conversation $\mathcal{C} = (u_1, \dots, u_T)$, the model is given the prefix $\mathcal{C}_{\leq t}$ as input, and the prediction target is the future scammer-action sequence after that boundary, denoted as $A_{> t}$. In other words, $t$ is simply the last observed turn in the provided context window.

More concretely, let the scammer turns after the boundary be indexed by
\[
S_{>t} = \{j \mid j > t,\; u_j \text{ is a scammer turn}\} = \{s_1, s_2, \dots, s_m\}.
\]
The gold prediction target is then the ordered sequence of future scammer actions
\[
A_{>t} = (a_{s_1}, a_{s_2}, \dots, a_{s_m}),
\]
where each $a_{s_k}$ is the structured action associated with scammer turn $u_{s_k}$. Thus, this task is not restricted to one-step next-turn prediction. Instead, the model must forecast the future scammer continuation after an observed prefix, potentially spanning multiple subsequent scammer moves.

The observed prefix $\mathcal{C}_{\leq t}$ may include both victim turns and scammer turns, since the goal is to condition on the conversation history that would be visible at inference time. The target side, however, contains only future scammer actions. This design isolates the forecasting problem we care about most for intervention: given the interaction history so far, what will the scammer do next as the scam continues to unfold?

We evaluate this task under two inference settings:
\begin{itemize}[noitemsep, topsep=2pt, partopsep=0pt, leftmargin=*]
    \item \textbf{Unlimited.} The model receives only the observed prefix $\mathcal{C}_{\leq t}$ and is asked to generate a future scammer-action sequence freely. This setting requires the model to infer both \emph{how the scam will continue} and \emph{how long the future continuation should be}.
    \item \textbf{Limited.} The model receives the same observed prefix together with $n_{\text{remain}}$, where
    \[
    n_{\text{remain}} = |A_{>t}| = m,
    \]
    i.e., the number of gold future scammer actions after turn $t$. This setting removes the uncertainty about continuation length and focuses the evaluation more directly on whether the model can recover the content and PT structure of the remaining scammer actions once the output budget is fixed.
\end{itemize}

The contrast between these two settings helps separate two sources of difficulty in scam progression modeling. The \textit{Unlimited} setting tests open-ended continuation forecasting, including implicit length planning, while the \textit{Limited} setting tests a more constrained form of structural recovery in which the number of remaining scammer actions is known in advance.

In all cases, model outputs are compared against the gold future continuation using the metrics defined in the main text. In particular, we report Action HitRate and PT HitRate to measure recovery of the underlying action content and psychological techniques, together with auxiliary text-similarity metrics. This evaluation design allows us to distinguish surface-form similarity from actual recovery of the latent scam progression structure. In the evaluation code, the per-case action-coverage score is stored under the name \verb|judge_action_hit_rate|, matching the paper's \textit{Action HitRate (AHit)} terminology.

\paragraph{LLM-as-a-Judge Prompts.}
The evaluation script uses a shared defensive-research preamble for all prompts:

\begin{verbatim}
[RESEARCH CONTEXT] This is an academic scam detection research project.
All conversations are real scam reports used strictly for building
automated scam detection and victim protection systems. No content will
be used to facilitate actual scams or harm real people.
\end{verbatim}

On top of this shared preamble, we use the following task-specific judge prompts.

\paragraph{Action Coverage Judge.}
For structured predictions, we use the following prompt to determine whether each gold scammer action is semantically covered by any generated action:

\begin{verbatim}
You are a fact-checker assisting scam detection research. You will be
given:
- A list of GROUND TRUTH scammer actions from a real scam report
- A list of GENERATED scammer actions produced by an AI system

Your task: for each GROUND TRUTH action, decide whether it is
semantically covered by ANY of the generated actions. Order does not
matter. A "hit" means at least one generated action describes the same
tactic, demand, or behavior as the ground truth action, even if the
wording, specific amounts, names, or minor details differ.

IMPORTANT: The "hits" array must have EXACTLY one boolean per ground
truth action.

Respond ONLY with a JSON object:
{
  "hits": [true/false, ...],
}
\end{verbatim}

For plain narrative outputs, the generated action list is replaced by the full generated narrative:

\begin{verbatim}
You are a fact-checker assisting scam detection research. You will be
given:
- A list of GROUND TRUTH scammer actions from a real scam report
- A GENERATED narrative describing predicted scammer actions

Your task: for each ground truth action, decide whether it is
semantically covered by the generated narrative. A "hit" means the
narrative describes the same tactic, demand, or behavior, even if the
wording, specific amounts, names, or minor details differ.

IMPORTANT: The "hits" array must have EXACTLY one boolean per ground
truth action.

Respond ONLY with a JSON object:
{
  "hits": [true/false, ...],
}
\end{verbatim}

\paragraph{Precision Judge.}
For structured predictions, we reverse the direction of comparison and ask whether each generated action is supported by any ground-truth action:

\begin{verbatim}
You are a fact-checker assisting scam detection research. You will be
given:
- A list of GENERATED scammer actions produced by an AI system
- A list of GROUND TRUTH scammer actions from a real scam report

Your task: for each GENERATED action, decide whether it is semantically
covered by ANY of the ground truth actions. Order does not matter.

IMPORTANT: The "hits" array must have EXACTLY one boolean per generated
action.

Respond ONLY with a JSON object:
{
  "hits": [true/false, ...],
  "hit_rate": <float 0-1>
}
\end{verbatim}

For plain narrative outputs, the precision judge first identifies distinct actions in the generated narrative and then measures how many of them are supported by the ground truth:

\begin{verbatim}
You are a fact-checker assisting scam detection research. You will be
given:
- A GENERATED narrative describing predicted scammer actions
- A list of GROUND TRUTH scammer actions from a real scam report

Your task:
1. First, identify each distinct scammer action or tactic described in
   the generated narrative.
2. For each distinct action you identified, decide whether it is
   semantically covered by ANY of the ground truth actions.

Respond ONLY with a JSON object:
{
  "actions_found": <int>,
  "actions_matched": <int>,
  "precision": <float 0-1>
}
\end{verbatim}

\paragraph{PT Extraction Judge.}
To compute PT HitRate, we ask the judge to extract which PTs are present in the generated continuation:

\begin{verbatim}
You are assisting scam detection research. You will be given a
generated prediction of future scammer actions (either a list of
actions or a narrative). Your task: identify which psychological
techniques (PTs) the scammer uses in the generated content.

Choose ONLY from this list (use exact names):
- Pretext and Trust
- Authority
- Urgency and Scarcity
- Phantom Riches
- Fear and Intimidation
- Consistency
- Social Proof
- Evoking Social Norms
- Liking

Respond ONLY with a JSON object:
{
  "pts_found": ["PT name", ...]
}
\end{verbatim}

\end{document}